%% file: main.tex
\documentclass{llncs}  

\usepackage{graphicx} 
\usepackage[T1]{fontenc}
\usepackage{amsmath}
\usepackage[english]{babel}
\usepackage{array,tabularx}
\usepackage{booktabs} 
\usepackage{makeidx}  
\usepackage{xparse}
\usepackage{moredefs}
\usepackage{lips}
\usepackage{epstopdf} 
\usepackage{color}
\usepackage{csquotes}
\usepackage{xcolor}
\usepackage{times}
\usepackage[hidelinks]{hyperref}
\usepackage[nolist,nohyperlinks]{acronym}
\usepackage{comment}
\usepackage{paralist}
\usepackage{cleveref}
\usepackage{wrapfig}
\usepackage{multirow}
\usepackage{microtype}
\usepackage{listings}
\usepackage[super]{nth}
\usepackage{natbib}
\usepackage{subcaption}

\usepackage[utf8]{inputenc} 
\usepackage[T1]{fontenc}    
\usepackage{hyperref}       
\usepackage{url}            
\usepackage{booktabs}       
\usepackage{amsfonts}       
\usepackage{nicefrac}       
\usepackage{microtype}      
\usepackage{xspace}
\usepackage{amsmath}
\usepackage{graphicx}
\usepackage{xcolor}
\usepackage{cleveref}

\newcommand\ie{i.\,e.,\xspace}

\newcommand{\E}{\mathbb{E}}

\usepackage{float}
\floatstyle{plaintop}
\restylefloat{table}

\usepackage[style=base,labelfont=bf,labelsep=period,tableposition=top,font=small]{caption}
\makeatletter
\renewcommand\@biblabel[1]{#1.}
\makeatother
\addto\captionsenglish{}

\usepackage{fancyhdr}
\fancypagestyle{WI_footer}{\fancyhf{}\fancyfoot[L]{\color{black}\scriptsize 18th International Conference on Wirtschaftsinformatik\\September 2023, Paderborn, Germany}}

\crefformat{footnote}{#2\footnotemark[#1]#3}
\expandafter\def\expandafter\UrlBreaks\expandafter{\UrlBreaks
  \do\a\do\b\do\c\do\d\do\e\do\f\do\g\do\h\do\i\do\j%
  \do\k\do\l\do\m\do\n\do\o\do\p\do\q\do\r\do\s\do\t%
  \do\u\do\v\do\w\do\x\do\y\do\z\do\A\do\B\do\C\do\D%
  \do\E\do\F\do\G\do\H\do\I\do\J\do\K\do\L\do\M\do\N%
  \do\O\do\P\do\Q\do\R\do\S\do\T\do\U\do\V\do\W\do\X%
  \do\Y\do\Z}

\newcolumntype{L}[1]{>{\raggedright\arraybackslash}p{#1}}   
\newcolumntype{C}[1]{>{\centering\arraybackslash}p{#1}}     
\newcolumntype{R}[1]{>{\raggedleft\arraybackslash}p{#1}}    

\begin{document}
\frontmatter          

\mainmatter              

\title{Improving the Efficiency of Human-in-the-Loop Systems: Adding Artificial to Human Experts}

\author{Johannes Jakubik \and
Daniel Weber \and
Patrick Hemmer \and
Michael Vössing\and
Gerhard Satzger} 

\institute{Karlsruhe Institute of Technology, Karlsruhe, Germany\\
\email{\{johannes.jakubik,daniel.weber,patrick.hemmer,michael.voessing,gerhard.satzger\}@kit.edu}}

\maketitle
\setcounter{footnote}{0}

\begin{abstract}
Information systems increasingly leverage artificial intelligence (AI) and machine learning (ML) to generate value from vast amounts of data. However, ML models are imperfect and can generate incorrect classifications. Hence, human-in-the-loop (HITL) extensions to ML models add a human review for instances that are difficult to classify. 
This study argues that continuously relying on human experts to handle difficult model classifications leads to a strong increase in human effort, which strains limited resources. To address this issue, we propose a hybrid system that creates artificial experts that learn to classify data instances from unknown classes previously reviewed by human experts. Our hybrid system assesses which artificial expert is suitable for classifying an instance from an unknown class and automatically assigns it. Over time, this reduces human effort and increases the efficiency of the system. Our experiments demonstrate that our approach outperforms traditional HITL systems for several benchmarks on image classification.

{\bfseries Keywords:} Human-in-the-Loop Systems, Artificial Experts, Human-AI Collaboration, Unknown Data.
\end{abstract}

\thispagestyle{WI_footer}


\section{Introduction}
\label{sec:introduction}

AI-based information systems have become increasingly prevalent and powerful over the last decade \citep[e.g.,][]{berente2021managing}. Yet, predictions (i.e., classification) of the incorporated machine learning (ML) models are subject to uncertainty and may be incorrect. As a result, human experts are often tasked with reviewing the predictions of these models to identify and override errors \citep{Dellermann2019,Gronsund2020,jakubik2022}. In these so-called human-in-the-loop (HITL) systems, a model might classify the majority of instances automatically. However, instances that are difficult for the model are assigned to a human expert for manual review. This concept makes HITL systems a viable means for human-AI collaboration and has promoted their applications in domains such as medicine \citep{boden2021,cai2019human} and manufacturing \citep{Cimini2020}. However, these systems typically still require a significant amount of human effort \citep{agnisarman2019survey}.

In HITL systems, detecting data from unknown classes to forward them to a human expert is essential to guarantee high levels of accuracy---especially in high-stake decision-making like the medical domain, finance, or autonomous driving \citep[][]{sanchez2022deep}. For example, when a model encounters an image of an ``unfamiliar'' traffic sign, it should treat it as unknown rather than attempting to match it to an existing class. However, this approach requires human experts to ``build'' knowledge regarding the novel class the instance belongs to. To address this issue and to reduce the required human effort, an approach is required that can incrementally learn from unknown data \citep{Keswani2022}.

In this work, we aim to improve the efficiency of HITL systems. We train additional ML models---called \textit{artificial experts}---that mimic the human expert's knowledge regarding classes that are unknown to the general ML model (i.e., the model trained to solve the task). 
Our so-called \textit{AI-in-the-Loop} (AIITL) system creates (i.e., trains) multiple artificial experts based on the knowledge incrementally acquired from human expert in traditional HITL systems. The AIITL system then allocates instances that are recognized as unknown to suitable artificial experts. To achieve this, we reinterpret out-of-distribution (OOD) detectors as an allocation mechanism. 

Each artificial expert is responsible for a separate set of classes unknown to the general model and uses an OOD detector to independently ``claim'' instances for classification that it considers to stem from one of the classes in their respective set. Data instances that the OOD detector deems to originate from an unknown class are rejected by the individual artificial expert. The human expert is only consulted when none or multiple artificial experts claim an instance for ``their'' set of classes. The human expert then provides his or her knowledge by assigning the instance to the correct artificial expert or instantiates a new one responsible for the novel class. The overall objective of the system is to reduce human effort while maintaining classification accuracy. Following \citet{Bansal2021}, we optimize a utility metric represented by the combination of classification accuracy and human effort. 

Overall, our contributions are as follows. First, we propose a novel technique for capturing human expert knowledge about unknown data instances and make it accessible by ``creating'' artificial experts. Second, we show that our approach outperforms traditional HITL systems by a large margin in terms of their utility---the combination of classification accuracy and human effort. Third, we reinterpret OOD detectors as allocation mechanisms for the collaboration of artificial and human experts---allowing for a coordination between artificial experts as ML models that can ``claim'' unknown instances. 
We provide our code at \url{https://github.com/jhnnsjkbk/AIITL}.
 
\section{Background and related work}
\label{section:related_work}

In the following, we review related work on HITL systems, the detection of data from unknown classes, and incremental learning.

\subsection{Human-in-the-Loop systems}
\label{subsec:hitl}

AI-based systems including a "human-in-the-loop" enable a machine learning model to consult a human expert for instances that are difficult to classify \citep{Dellermann2019a,Gronsund2020,jakubik2022}. For example, in the medical domain, it is essential that machine learning models forward x-ray images that are difficult to classify (i.e., the model is uncertain about the prediction) to physicians for manual inspection.  
In the literature, several setups exist in which human experts augment and complement ML models: HITL systems are employed in supervised learning \citep[e.g.,][]{wang2016human,Kamar2016,Wu2021}, semi-supervised learning \citep[e.g.,][]{wrede2019smart,weber2021better}, and reinforcement learning \citep[e.g.,][]{wu2022toward,liu2019deep,liang2017human,elmalaki2021fair}. 
However, these approaches generally require repetitive human effort that is growing with the number of unknown instances and the inaccuracy in detecting such instances. The resulting strain placed on human resources often renders HITL systems inefficient (see \citeauthor{agnisarman2019survey} \citeyear{agnisarman2019survey}). Therefore, any approach that can reduce human effort required in these systems is highly desirable. 

A literature search identified two recent works pursuing this objective, however with slightly different approaches: First, \cite{Keswani2022} consider a HITL system where the classifier is incrementally re-trained on novel, unseen data. For this re-training, the ML model acquires human expert knowledge for instances with low model confidence. Thus, the overall goal is to incrementally mimic several human experts with a \textit{single ML model} in order to reduce subsequent effort. 
In contrast, we propose various seperate artificial experts that individually learn to classify instances from specific domains and collaboratively classify novel instances in order to reduce human effort.
Unlike the approach of \cite{Keswani2022}, our technique does not require to train an additional deferral model for the collaboration of multiple agents in the system. 
Second, \cite{Wu2021_DLA} have recently proposed the collaboration between two ML models in an HITL system. Their approach defers instances to human experts based on the alignment of these ML models as part of a multi-model collaboration. However, their work is specifically tailored to document layout analysis, while we propose a system to reduce the effort in HITL systems in general.

\subsection{Detection of data instances from unknown classes}
\label{subsec:ood}

In our work, we make use of out-of-distribution detectors from the field of computer science and reinterpret them as allocation mechanisms. These OOD detectors allow to determine whether a data instance originates from an unknown class. The differentiation of known and unknown data is achieved based on whether unknown data originates from the same underlying distribution as known data or not. For example, everyday images of cars follow the same (or very similar distribution) but these images follow a different distribution than x-ray images from the medical domain. This approach can also be applied to distinguish more similar classes. 
In our case, we will consider classes from one dataset to be known, while classes from a range of other datasets are unknown and incrementally learned over time. 
In general, OOD detection aims at identifying data from unknown distributions \citep{Liang2018, Lee2018,Hsu2020,Liu2020_EnergyOOD}). 
Among the most widespread approaches for OOD detection are \textit{ODIN} \citep{Liang2018} and \textit{Mahalanobis-based OOD} detection \citep{Lee2018}. These two approaches are popular benchmarks utilized in recent studies \citep{Hsu2020, Liu2020_OOD, Mohseni2020}. In the context of HITL systems, OOD detection is typically considered during the training in active learning and few-shot learning \citep{Han2021, Liu2020_OOD, Wan2021} or for explainability \citep{bansal2018}. 
Importantly, current literature finds a lack of handling detected unknown data in general \citep{Geng2021}. We later address this by using OOD detectors as allocation mechanisms and by processing detected data from unknown classes within our AIITL system.

\subsection{Incremental learning}

Deep neural networks face severe difficulties when learning from evolving streams of training data. This phenomenon is often referred to as \textit{catastrophic forgetting} \citep{mittal2021essentials,li2017learning,rebuffi2017icarl} as---while adapting to the new classes---the performance of the ML model on the original classes deteriorates. That means the accuracy of ML models drops strongly, when the model is incrementally trained on new data.
In the context of reducing human effort in HITL systems, this implies that we cannot simply fine-tune our model incrementally based on acquired human expert knowledge without facing catastrophic forgetting. Retraining the model on the entire set of known and unknown data each time unknown instances emerge would result in high computational cost and, thus, does not represent a suitable alternative either. 

To tackle catastrophic forgetting, class-incremental learning aims at making models more robust against new data that becomes available over time. Researchers have proposed approaches such as exemplar selection \citep[e.g.,][]{castro2018end,hou2019learning,wu2019large,iscen2020memory,liu2020mnemonics}, forgetting-constraint \citep[e.g.,][]{li2017learning,rebuffi2017icarl,hou2019learning,yu2020semantic}, and bias removal methods \citep[e.g.,][]{hou2019learning,wu2019large,zhao2020maintaining}. 
One of the most popular approaches in incremental learning is Deep Model Consolidation \citep{zhang2020class}. The approach consists of a general model for old (\ie known) data and a separate model that is responsible for the new classes. We later follow a similar idea by generating separate models for novel data from unknown classes to reduce the human effort in HITL systems. 

\section{Methodology}
\label{sec:aiitl}

\begin{figure}[ht]
\begin{center}
\includegraphics[width=\textwidth]{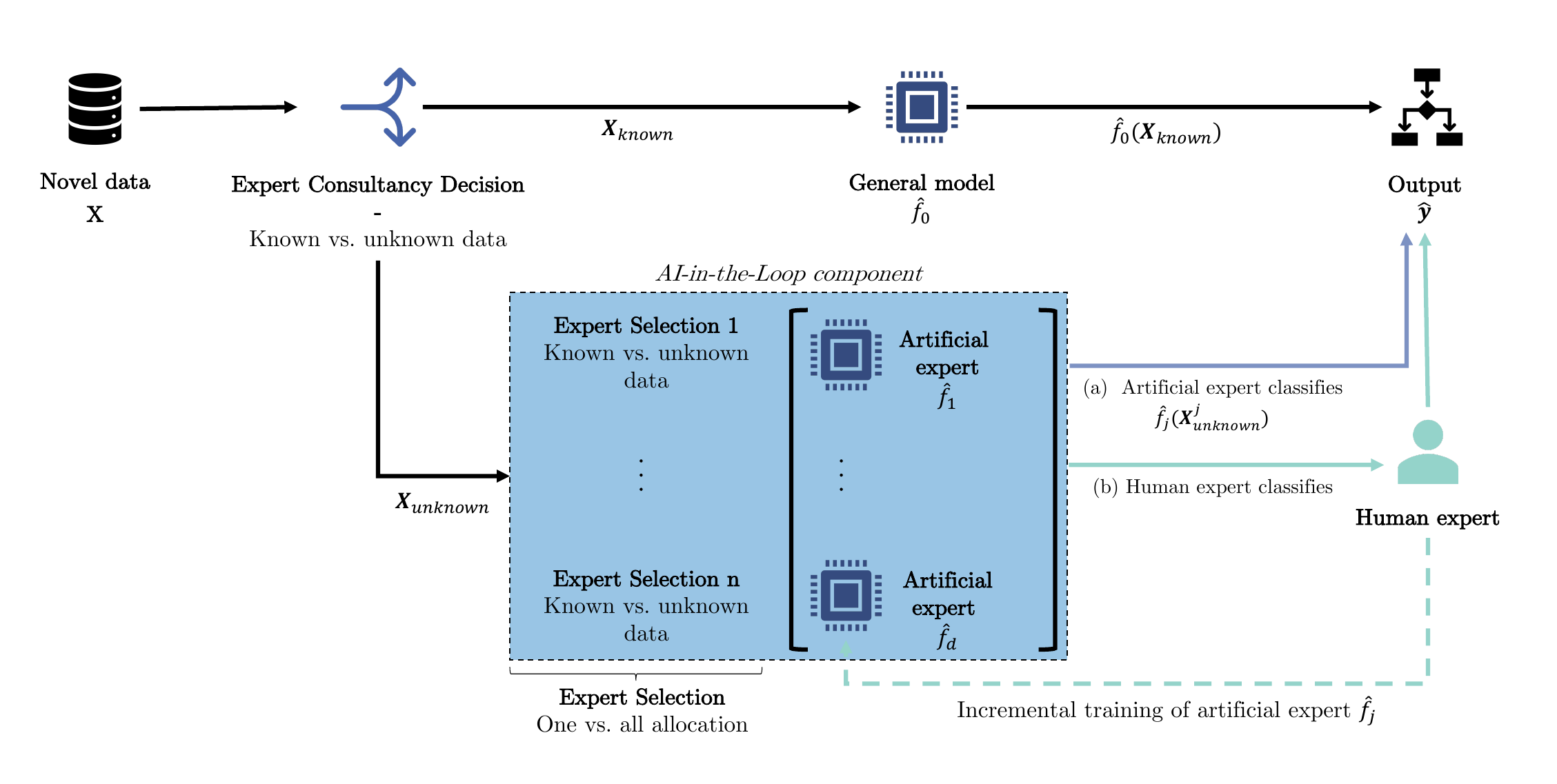}
\caption{Illustration of the AIITL system with artificial experts to support the human expert of a Human-in-the-Loop system.}
\label{system_design}
\end{center}
\end{figure}

The task of the AIITL system is to classify data from both known and unknown classes. We train a \textit{general} model on data from the known classes. Unknown data originates from a set of unknown classes. 
The classification of an instance is then either conducted by the general model, one of $n$ artificial experts, or a human expert. Our objective is to improve the utility of the system that is influenced by the level of human effort and the overall classification accuracy. The hybrid system is detailed in Figure~\ref{system_design} and will be explained in the following. 

\subsection{Notation}
\label{subsec:problem_formalization}

The objective of the AIITL system is to reduce repetitive manual reviewing effort of human experts. For this, artificial experts need to mimic human experts on data instances that are unknown for the general model. 
In line with our research questions, we developed a system of human and artificial experts that aid a general ML model in classifying images. For this, we design artificial experts as an optional support that actively try to classify instances---and only forward instances to a human expert that both the general model and none of the artificial experts could classify. 
We formalize our approach in the following. 
Let $\mathcal{X} \in \mathbb{R}^{N}$ and $\mathcal{Y} \in \mathbb{N}$ denote the input and target space of the AIITL system. The input space consists of the images that need to be classified by our system, while the target space refers to the classication outcomes for each of the input images.
In our approach, we refer to known data by $\mathbf{X}_{known}$ and define data from unknown classes as $\mathbf{X}_{unknown}$.
The overall task of the AIITL system is to provide a set of classifications $\hat{\mathbf{y}}$ for the input data instances carried out by either the general ML model, a maximum of $n$ supplementary artificial experts, or a human expert. 
Each of the ML models (\ie the general ML model and the artificial experts) are iteratively trained to classify data from a specific domain. 
This results in individual classification functions
\begin{equation}
f_p: \mathcal{X}^j \rightarrow \mathcal{Y}^{j}, \quad \mathbf{X}^j \mapsto \mathbf{y}^j
\label{F_j}
\end{equation}
for each of the ML models in the AIITL system, where $j$ refers to the index of the model. Note, that these functions are unknown a priori and are approximated by convolutional neural networks as $\hat{f}_p(\mathbf{X}^j) = \hat{\mathbf{y}}^j$ when training the AIITL system.

Following \citet{Bansal2021}, we use a utility score that combines the system's accuracy $\phi(\mathbf{X}, \hat{f})$ and the human effort $\rho(\mathbf{X}, \hat{f})$ to approximate the efficiency of the AIITL system for our evaluation. It is defined as:
\begin{equation}
U(\mathbf{X}, \hat{f}) = \alpha \cdot \phi(\mathbf{X}, \hat{f})  - \beta \cdot \rho(\mathbf{X}, \hat{f})
\label{utility}
\end{equation}
with $\mathbf{X} = \bigcup_j \mathbf{X}^j$ and $\hat{f} = \bigcup_p \hat{f}_p$. Accuracy quantifies the classification quality of the AIITL system. Human effort refers to the number of instances that are classified by the human expert in relation to the total number of instances. 
For the calculation of the utility score, we follow \citet{Bansal2021} and set $\alpha = 1$ and $\beta = 0.5$. That means, that we consider costs of inaccuracies to be two times higher than cost of manual review by human experts. In our experiments, we later demonstrate that our results are robust against changes in these parameters.

\subsection{Allocation mechanisms}
\label{subsec:allo_mech}

We implement two allocation stages that guide instances through the AIITL system. The \textit{Expert Consultancy Decision} is part of the traditional HITL system, deciding whether an instance is known or unknown utilizing OOD detection. Instances from known classes are then classified by the general model, while unknown instances are passed to an expert for review. The \textit{Expert Selection} allows us to defer detected unknown data to artificial experts that are incrementally trained using data and labels from the manual review of the human expert. 
For the Expert Selection, we reinterpret OOD detectors as allocation mechanisms. To this end, we fit an OOD detector for each of the artificial experts. 
The OOD detectors then claim an instance for the respective expert if the instance is considered to originate from a known class (\ie the instance is drawn from the experts' training distribution). 
If an instance is claimed by none or multiple artificial experts, the instance is allocated to a human expert for manual review. In the following, we briefly introduce the employed allocation mechanisms:

\begin{itemize}
    \item \textbf{ODIN} is an OOD detector based on temperature scaling and perturbations of the input data \citep{Liang2018}. Model- and domain-specific scores are compared to 
    thresholds. Based on this score, an instance is either defined as originating from a known class or from an unknown class. 
    \item \textbf{Mahalanobis-based OOD (MAHA)} computes scores based on input perturbation and the Mahalanobis-distance between the input and the closest class-conditional Gaussian distribution \citep{Lee2018}. These scores are leveraged to classify an instance to part of the set of known classes or stem from an unknown class.
    \item \textbf{Gating model} is a separate model from the field of so-called ``Mixture of Experts'' that is trained to allocate instances to the artificial experts \citep{Jacobs1991}. In contrast to the OOD detectors, this model is explicitly trained on allocating data instances to artificial experts.
\end{itemize}

\section{Experiments}
\label{sec:experiments}

In the following, we describe our experimental setup. For the evaluation, we use popular datasets for image classification (e.g., CIFAR-10, SVHN). 
We evaluate the AIITL system under incremental data availability over $30$ discrete steps. After training the general model, known and unknown data incrementally appears over the 30 steps. We train the ``candidate'' artificial experts on the labels that are iteratively generated by human experts during manual reviewing. In our implementations, we simulate a single human expert and assume a perfect classification accuracy for all input instances following related literature \citep[e.g.,][]{Bansal2021}. We evaluate the performance of the AIITL system in a separate test batch (\ie after the 30 steps of incremental learning).

We use a small proportion of the data from manual reviewing for validation and test of the artificial experts. 
We split the remaining data in 80\% data for training and 20\% for test.
We then utilize the accuracy of artificial experts on the test set as indicator when a specific artificial experts should be included in the AIITL system. This is in line with \cite{Bansal2021}, who utilize the model confidence, but more directly related to the utility metric. When the test accuracy of an artificial expert exceeds a threshold of 95\%, the system includes the artificial expert and considers it during the allocation of detected unknown data. 
In Section \ref{sec:robustness}, we conduct sensitivity analyses on this threshold.

\subsection{Benchmarks}
Our evaluation procedure uses three benchmarks. First, we employ the general model from the HITL system without the option of a manual review from the human experts. This benchmark can be considered as a full automation baseline and prevents human effort but is likely to have a low classification accuracy in the presence of unknown data. Second, we utilize a traditional HITL system, where the general model classifies known instances and the human expert is consulted for manual reviewing of the unknown instances. 
This HITL system also constitutes the initial version of each AIITL system (\ie step~1) and allows us to assess the merit of introducing artificial experts over the steps. Finally, we compare the AIITL system with the HITL system under \textit{perfect} allocation of unknown data, which is an upper bound in terms of classification accuracy.

\subsection{Datasets} For our experiments, we build a dataset out of four well-known image classification datasets. While CIFAR-10 \citep{CIFAR-10} is used to train the general model and, thus, represents the known classes, three others represent unknown data classes: SVHN \citep{SVHN}, MNIST \citep{MNIST} and Fashion-MNIST \citep{Fashion-MNIST}.
CIFAR-10 consists of 60,000 images of ten different classes (e.g., airplane, car, truck). Fashion-MNIST contains 60,000 images of Zalando's articles from ten different classes. MNIST consists of 60,000 images of handwritten digits and SVHN includes 600,000 images of printed digits from pictures of house number plates from Google Street View. We utilize the predefined train-test splits for the datasets. 

\subsection{Training} We employ a Wide-ResNet-28-10 architecture \citep{Zagoruyko2016} for the general model and a total of three artificial experts, where each artificial expert will be responsible for a separate set of unknown data. For the gating model, we make use of a DenseNet-121 \citep{Huang2016} pretrained on ImageNet \citep{Russakovsky2014}. We train the general model for 200 epochs with a batch size of 256 and SGD as the optimizer with a learning rate of 0.1. The artificial experts are iteratively trained for 300 epochs with a 128 batch size using the SGD optimizer with a learning rate of 0.1. All models were trained on a NVIDIA A100 GPU.

\subsection{Results} 
\begin{figure}[ht]
\begin{center}
\includegraphics[width=0.75\textwidth]{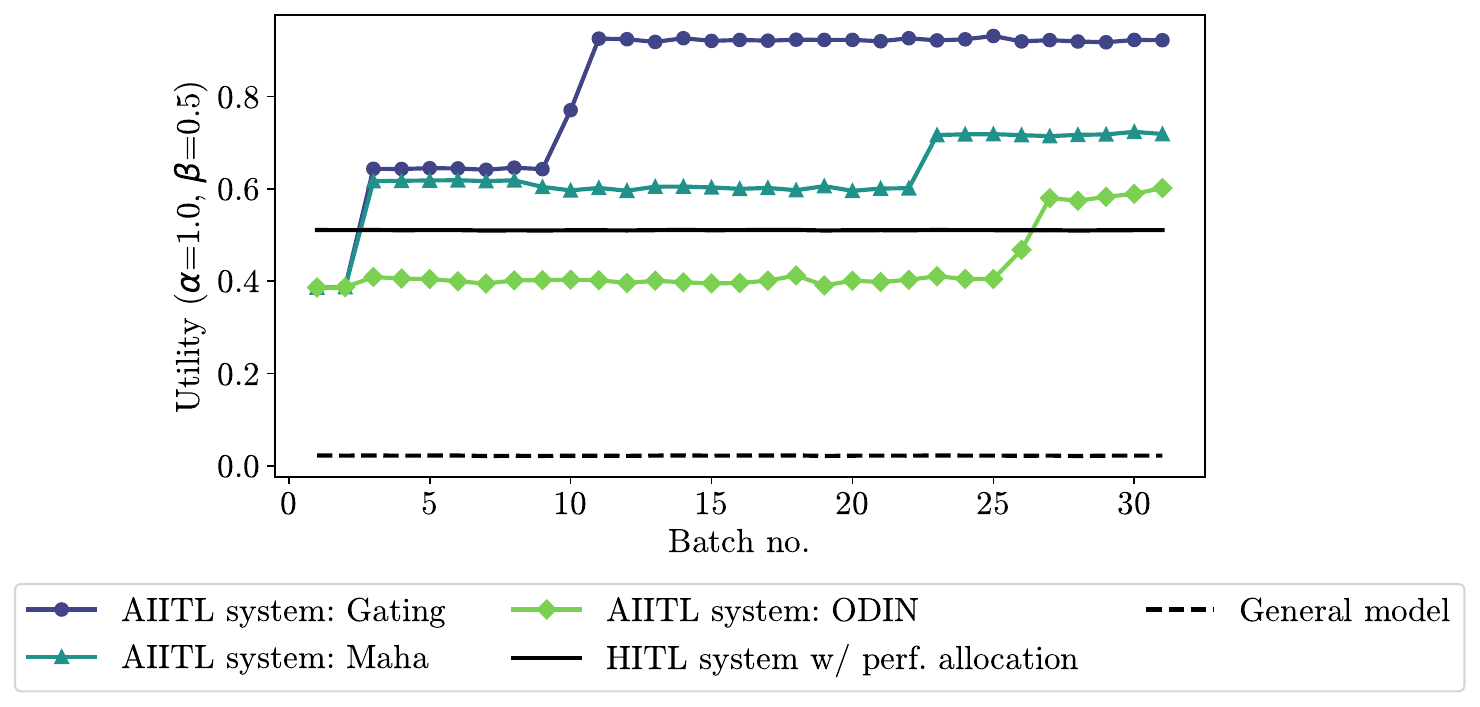}
\caption{Utility score of the AIITL system for different allocation mechanisms and baselines.}
\label{bench_multi_dyn_utility0.5}
\end{center}
\end{figure}

We present the utility scores of the AIITL system and the proposed baselines in Figure \ref{bench_multi_dyn_utility0.5}. After 30 steps of incremental learning, we observe that the proposed hybrid system outperforms the \textit{perfect} HITL system with all allocation mechanisms. While the performance of the HITL system under perfect allocation results in an utility of 0.51, our system achieves scores of 0.92 based on the allocation using the gating model, 0.73 with allocation based on the Mahalanobis method, and 0.61 for allocation based on the ODIN method.
This suggests that even when we assume a perfect allocation of known and unknown images (i.e., the general model only classifies known images and the human expert only reviews unknown images), the HITL system is less efficient than our hybrid system.

Table \ref{benchmark_multi_dynamic_acc_cov} reports the accuracy and human effort of the initial HITL system from step 1 (i.e., a traditional HITL system with imperfect allocation using allocation mechanisms) and the resulting AIITL system after 30 steps. 
The AIITL system outperforms the HITL system (\ie in the absence of artificial experts) both in terms of accuracy and human effort for allocation based on the gating model and MAHA. The accuracy increases by 17pp from 75\% to 92\%. The human effort is reduced by 73pp and 34pp by the gating model and MAHA, respectively. For the allocation based on ODIN the accuracy slightly declines from 75\% to 74\%, while the human effort is reduced significantly from 0.73 to 0.27. Overall, this demonstrates that the suggested hybrid system improves the efficiency strongly compared to HITL systems. 

\input{Tables/bench_multi_dynamic_acc_cov}

\subsection{Analyzing the influence of varying weights for human effort on the system efficiency}\label{sec:robustness}

\begin{figure}[H]
     \centering
     \begin{subfigure}[c]{0.024\textwidth}
         \vspace{-4.8cm}
         \centering
         \includegraphics[width=1\textwidth]{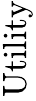}
     \end{subfigure}
     \vspace{0.2cm}
     \begin{subfigure}[b]{0.3\textwidth}
         \centering
         \includegraphics[width=.935\textwidth]{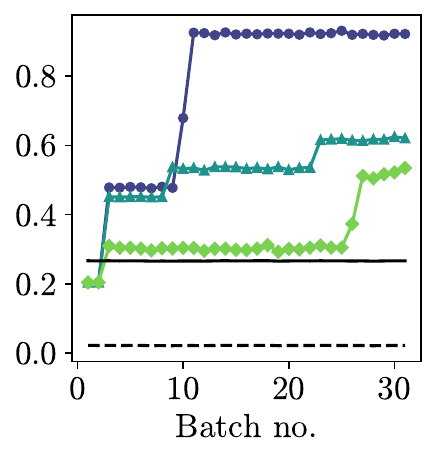}
         \caption{$\beta=0.75$}
         \label{benchmark_multi_beta1.0}
     \end{subfigure}
     \vspace{0.2cm}
     \begin{subfigure}[b]{0.3\textwidth}
         \centering
         \includegraphics[width=.935\textwidth]{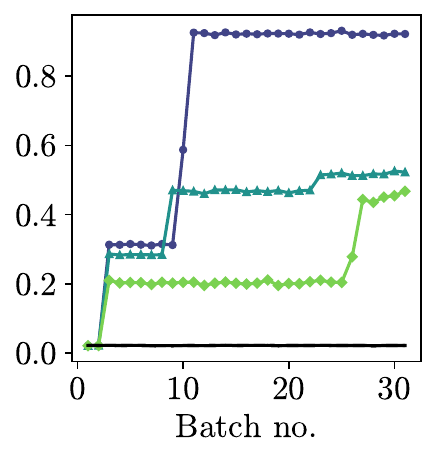}
         \caption{$\beta=1.0$}
         \label{benchmark_multi_beta1.5}
     \end{subfigure}
     \vspace{0.2cm}
     \begin{subfigure}[b]{0.3\textwidth}
         \centering
         \includegraphics[width=.935\textwidth]{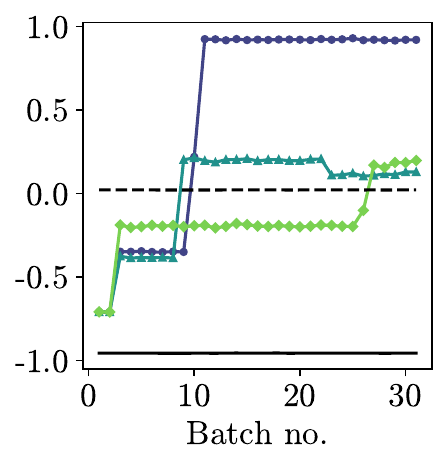}
         \caption{$\beta=2.0$}
         \label{benchmark_multi_beta2}
     \end{subfigure}
     \vspace{0.2cm}
     \begin{subfigure}[b]{1\textwidth}
         \centering
         \includegraphics[width=0.75\textwidth]{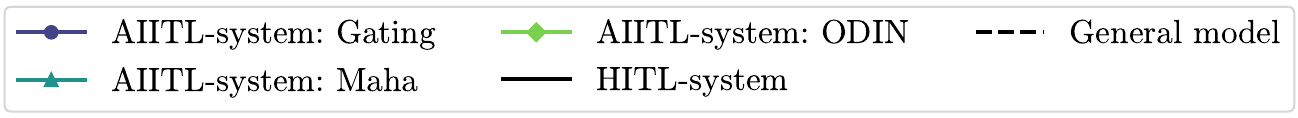}
     \end{subfigure}
        \caption{Sensitivity analysis regarding the influence of the weight of human effort $\beta$ on the overall utility of the AIITL system and the baselines ($\alpha=1$).}
        \label{fig:benchmark_multi_beta_sensitivityanalysis}
\end{figure}

We conduct sensitivity analyses to better comprehend the influence of the weight of human effort in Eq.~\ref{utility}. 
For that, we vary the weight of human effort and depict examples in Figure~\ref{fig:benchmark_multi_beta_sensitivityanalysis}. Across experiments, we find consistent improvements of our hybrid system over traditional HITL systems. In addition, we evaluate when traditional HITL systems become preferable in our experiments and found that HITL systems improve over our hybrid system only when human effort is more than ten times less important than accuracy (\ie $\beta < 0.1$).
In our experiments, AIITL systems consistently outperform HITL systems when accuracy is less than ten times more important than human effort (\ie $\beta > 0.1$).
Overall, our results demonstrate strong increases in the efficiency of the proposed hybrid AIITL system compared to traditional HITL systems.

\subsection{Applying the hybrid system in the context of semantically similar classes}

\begin{figure}[H]
\begin{center}
\includegraphics[width=1\textwidth]{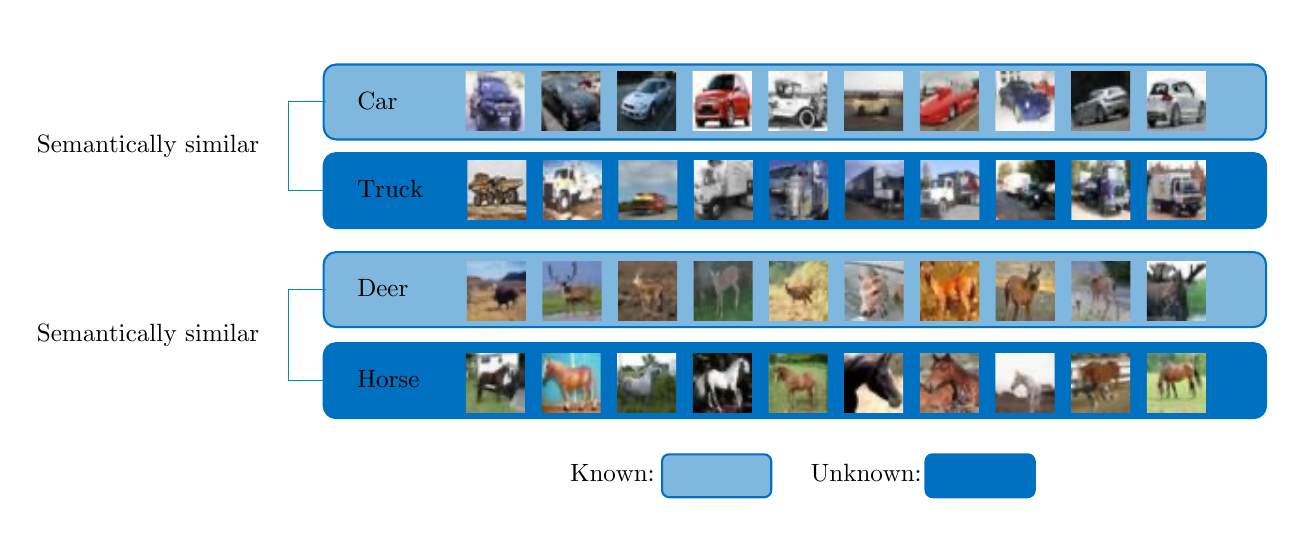}
\caption{Visual examples of semantically similar classes that are considered as known and unknown classes respectively. The general model is trained on known classes, while the artificial experts learn to classify unknown classes based on incremental feedback from the human expert.}
\label{fig:similiar_classes}
\end{center}
\end{figure}

In the previous section, we focused on unknown classes that originate from other datasets than the training dataset for the general model (e.g., the general model was trained on CIFAR-10 images, while unknown data originates from SVHN). In this section, we show that our hybrid system improves the efficiency of HITL systems even when we sample a range of classes of the training set to be known, while other classes of this dataset are treated as unknown. For this, we select the first six classes of CIFAR-10 to be known, while we consider the remaining four as unknown classes. For example, the classes ``car'' and ``deer'' represent known classes, while the semantically similar classes ``truck'' and ``horse'' represents unknown classes (see Figure~\ref{fig:similiar_classes}).

\begin{figure}[H]
\begin{center}
\includegraphics[width=0.75\textwidth]{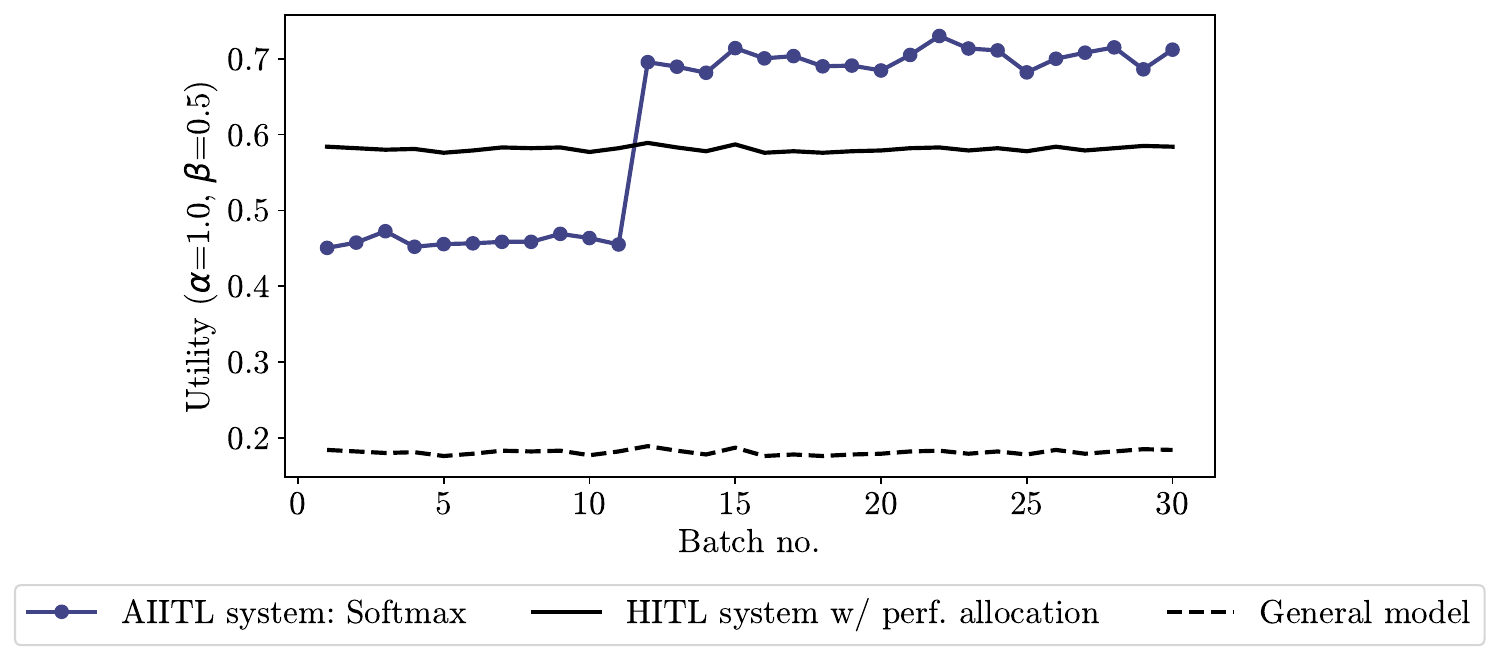}
\caption{Utility score of our hybrid system and baselines in a setting where unknown classes originate from the same dataset as known classes. Graphs depict performances on the CIFAR-10 dataset, where six classes are considered to be known, while the remaining four classes are treated as unknown.}
\label{benchmark_single_dyn_utility0.75}
\end{center}
\end{figure}

We then train the general model on the known classes, while a single artificial expert and the human expert are later responsible for classifying the four unknown classes. Following common practice for detecting unknown classes that are part of the same set as known classes, we utilize softmax thresholding as a simple baseline to detect unknown instances \citep{Geng2021}. We present our results in Figure \ref{benchmark_single_dyn_utility0.75}. Overall, we observe that our hybrid system outperforms the traditional HITL system, the HITL system under perfect allocation, and the general model by a large margin. 
Our results are in line with the previous findings and demonstrate that the AIITL system can be leveraged in contexts where known and unknown classes are very similar.

\section{Discussion}
\label{sec:discussion}

Our work results in a range of implications for research and practice in the field of HITL systems and hybrid systems that leverage both ML and human expert knowledge. 

First, we observe inefficiencies in traditional HITL systems that are created by allocation of tasks to a human experts that are difficult to classify for the ML model. This results in substantial human effort that ultimately can inhibit the application of HITL systems in general. We demonstrate that a hybrid team of artificial and human experts can significantly reduce the human effort and thereby increase the efficiency of the overall system. For practitioners, this implies that knowledge of domain experts can continuously be ingested into artificial experts (i.e., ML models) that can support human experts on repetitive tasks. This frees human capacities for creative, and more complex tasks. For research, our approach demonstrates that a hybrid team of human and artificial experts can successfully support a general ML model in the presence of data that originates from unknown classes. Especially in real-world applications, well-defined processes to handle data from unknown classes is essential to ensure a required level of system accuracy.

Second, we observe two interesting trade-offs in the design of our hybrid system that need to be taken account for when utilizing our approach in real-world applications: (a) the choice of the allocation mechanism and (b) the required accuracy level of artificial experts to include them in the hybrid team. For the choice of the allocation mechanism, we observe a trade-off between the efficiency level (i.e., utility score) and the capability of adapting to data from unknown classes. For example, the gating model achieves a very strong utility, while it is limited in the ability to detect data from novel classes. The human expert first needs to feed the gating model with a sufficient number of instances from the encountered unknown class so that the gating model is able to allocate instances from that class correctly. On the other hand, ODIN and MAHA techniques achieve slightly lower performances (still outperforming traditional HITL systems by a large margin), but are adapting more quickly to a novel set of unknown classes. When designing hybrid systems in practice, it is critical to carefully select a suitable allocation mechanism. The second trade-off refers to the accuracy of artificial experts that is required to include the expert in the hybrid system. Generally, the higher the accuracy of an artificial expert, the better the overall performance of the system. However, achieving an increased performance of the artificial experts typically requires a higher amount of labeled data and, therefore, of manual reviews by the human expert. Defining the required level of accuracy of the artificial expert in applications in research and practice will require an in depth understanding of the costs of misclassifications (i.e., inaccuracies) and the costs of manual reviews by the human expert.

Third, we see exciting parallels between our hybrid system and the field of human-AI collaboration as well as complementary team performance. In the latter, human and AI achieve a joint accuracy that none of them would have achieved individually. These approaches focus on maximizing the accuracy, while we add an additional dimension by focusing on the human effort. Coupling these approaches may result in even more promising performances in terms of the overall efficiency of the system.

As any research, ours is not free of limitations. While we evaluated our approach on a wide range of image classification datasets, evidence from the field of, for example, natural language and from structured data is missing. Moreover, the approach needs to be validated in real-world situations with domain experts representing the ``human-in-the-loop'' to assess actual cost savings and efficiency improvements. Thus, as a next step, we want to explore variations of AIITL systems in different use cases and on real-world datasets. Finally, while our approach is general, we evaluated our approach in a linear setting due to computational limitations, meaning that the number of artificial experts linearly increased with the number of unknown domains. It will be interesting for future research to investigate how e.g., a constant number of artificial experts performs in settings with an increasing number of unknown domains. Thus, additional research on the optimal number of artificial experts is necessary.
%
To increase the practical applicability, we are interested in exploring additional mechanisms to assess when an artificial expert should be included in the system. 

\section{Conclusion}
\label{sec:conclusion}

In this work, we argue that consulting human experts for manual review of difficult model classifications leads to a strong increase in human effort. Retraining an ML model on incrementally appearing novel classes is not an alternative as it leads to a significantly declining performance on the base classes---a phenomenon called ``catastrophic forgetting''. To improve the efficiency of HITL systems, we introduce several artificial experts that learn to classify data from unknown classes based on the manual review from human experts. The resulting hybrid system of human and artificial experts improves the efficiency of HITL systems by reducing human effort and is robust against catastropic forgetting. We additionally introduce allocation mechanisms within our hybrid system that allow us to automatically assign detected unknown data to a suitable artificial expert. 
Overall, we find that our hybrid system outperforms traditional HITL systems by a large margin in terms of their utility---the combination of classification accuracy and human effort---across a range of benchmarks. 


\bibliographystyle{agsm}
\bibliography{literature.bib}

\end{document}

%% file: Tables/bench_multi_dynamic_acc_cov.tex





\begin{table}[ht]
\caption{Overview of accuracy and human effort of the AIITL system after 30 steps compared to the performance of a traditional HITL system. Best performances are highlighted in bold.}

\centering

\scalebox{0.9}{

\begin{tabular}{lccccc}
\toprule
\textbf{} & {HITL} & \multicolumn{3}{c}{{AIITL}} \\\cmidrule{3-5}
 &      &     Gating & MAHA & ODIN                 \\
\midrule
{Human effort $\rho(\mathbf{X}, \hat{f})$} &  0.73 &   
\textbf{0.00} & 0.39 & 0.27 \\
{Accuracy $\phi(\mathbf{X}, \hat{f})$       } &  0.75 & \textbf{0.92} & \textbf{0.92} & 0.74  \\\midrule
{Utility $U(\mathbf{X}, \hat{f})$} &  0.39 & \textbf{0.92} & 0.73 & 0.61  \\
\bottomrule
\end{tabular}}

\label{benchmark_multi_dynamic_acc_cov}
\end{table}

%% file: main.bib
@inproceedings{Bansal2021,
   author = {Gagan Bansal and Besmira Nushi and Ece Kamar and Eric Horvitz and Daniel S. Weld},
   booktitle = {Proceedings of the Thirty-Fifth AAAI Conference on Artificial Intelligence (AAAI-21)},
   publisher={AAAI Press},
   volume={35},
   pages={11405--11414},
   title = {{Is the Most Accurate AI the Best Teammate? Optimizing AI for Teamwork}},
   year = {2021},
}

@inproceedings{bansal2018,
   author = {Gagan Bansal and Daniel S Weld and Paul G Allen},
   year = {2018},
   title = {{A Coverage-Based Utility Model for Identifying Unknown Unknowns}},
   booktitle = {Proceedings of the Thirty-Second AAAI Conference on Artificial Intelligence (AAAI-18)},
   publisher={AAAI Press},
   volume={32},
   pages={},
}

@inproceedings{zhao2020maintaining,
  title={Maintaining discrimination and fairness in class incremental learning},
  author={Zhao, Bowen and Xiao, Xi and Gan, Guojun and Zhang, Bin and Xia, Shu-Tao},
  booktitle={Proceedings of the IEEE/CVF Conference on Computer Vision and Pattern Recognition},
  pages={13208--13217},
  year={2020}
}

@inproceedings{wu2019large,
  title={Large scale incremental learning},
  author={Wu, Yue and Chen, Yinpeng and Wang, Lijuan and Ye, Yuancheng and Liu, Zicheng and Guo, Yandong and Fu, Yun},
  booktitle={Proceedings of the IEEE/CVF Conference on Computer Vision and Pattern Recognition},
  pages={374--382},
  year={2019}
}

@inproceedings{yu2020semantic,
  title={Semantic drift compensation for class-incremental learning},
  author={Yu, Lu and Twardowski, Bartlomiej and Liu, Xialei and Herranz, Luis and Wang, Kai and Cheng, Yongmei and Jui, Shangling and Weijer, Joost van de},
  booktitle={Proceedings of the IEEE/CVF Conference on Computer Vision and Pattern Recognition},
  pages={6982--6991},
  year={2020}
}

@inproceedings{hou2019learning,
  title={Learning a unified classifier incrementally via rebalancing},
  author={Hou, Saihui and Pan, Xinyu and Loy, Chen Change and Wang, Zilei and Lin, Dahua},
  booktitle={Proceedings of the IEEE/CVF Conference on Computer Vision and Pattern Recognition},
  pages={831--839},
  year={2019}
}

@inproceedings{rebuffi2017icarl,
  title={{iCaRL: Incremental classifier and representation learning}},
  author={Rebuffi, Sylvestre-Alvise and Kolesnikov, Alexander and Sperl, Georg and Lampert, Christoph H},
  booktitle={Proceedings of the IEEE conference on Computer Vision and Pattern Recognition},
  pages={2001--2010},
  year={2017}
}

@article{li2017learning,
  title={Learning without forgetting},
  author={Li, Zhizhong and Hoiem, Derek},
  journal={IEEE transactions on pattern analysis and machine intelligence},
  volume={40},
  number={12},
  pages={2935--2947},
  year={2017},
  publisher={IEEE}
}

@inproceedings{iscen2020memory,
  title={Memory-efficient incremental learning through feature adaptation},
  author={Iscen, Ahmet and Zhang, Jeffrey and Lazebnik, Svetlana and Schmid, Cordelia},
  booktitle={European conference on computer vision},
  pages={699--715},
  year={2020},
  organization={Springer}
}

@inproceedings{liu2020mnemonics,
  title={{Mnemonics training: Multi-class incremental learning without forgetting}},
  author={Liu, Yaoyao and Su, Yuting and Liu, An-An and Schiele, Bernt and Sun, Qianru},
  booktitle={Proceedings of the IEEE/CVF conference on Computer Vision and Pattern Recognition},
  pages={12245--12254},
  year={2020}
}

@inproceedings{castro2018end,
  title={End-to-end incremental learning},
  author={Castro, Francisco M and Mar{\'\i}n-Jim{\'e}nez, Manuel J and Guil, Nicol{\'a}s and Schmid, Cordelia and Alahari, Karteek},
  booktitle={Proceedings of the European conference on computer vision (ECCV)},
  pages={233--248},
  year={2018}
}

@inproceedings{mittal2021essentials,
  title={Essentials for class incremental learning},
  author={Mittal, Sudhanshu and Galesso, Silvio and Brox, Thomas},
  booktitle={Proceedings of the IEEE/CVF Conference on Computer Vision and Pattern Recognition},
  pages={3513--3522},
  year={2021}
}

@inproceedings{zhang2020class,
  title={Class-incremental learning via deep model consolidation},
  author={Zhang, Junting and Zhang, Jie and Ghosh, Shalini and Li, Dawei and Tasci, Serafettin and Heck, Larry and Zhang, Heming and Kuo, C-C Jay},
  booktitle={Proceedings of the IEEE/CVF Winter Conference on Applications of Computer Vision},
  pages={1131--1140},
  year={2020}
}

@article{boden2021,
   author = {Anna C.S. Bod\'en and Jesper Molin and Stina Garvin and Rebecca A. West and Claes Lundstr\"om and Darren Treanor},
   number = {2},
   journal = {Histopathology},
   pages = {210--218},
   publisher = {John Wiley & Sons},
   title = {{The human-in-the-loop: An evaluation of pathologists' interaction with artificial intelligence in clinical practice}},
   volume = {79},
   year = {2021},
}

@article{Cimini2020,
   author = {Chiara Cimini and Fabiana Pirola and Roberto Pinto and Sergio Cavalieri},
   journal = {Journal of Manufacturing Systems},
   pages = {258--271},
   publisher = {Elsevier B.V.},
   title = {{A human-in-the-loop manufacturing control architecture for the next generation of production systems}},
   volume = {54},
   number={},
   year = {2020},
}

@article{agnisarman2019survey,
  title={{A Survey of Automation-Enabled Human-In-The-Loop Systems For Infrastructure Visual Inspection}},
  author={Agnisarman, Sruthy and Lopes, Snowil and Madathil, Kapil Chalil and Piratla, Kalyan and Gramopadhye, Anand},
  journal={Automation in Construction},
  volume={97},
  pages={52--76},
  year={2019},
  publisher={Elsevier}
}

@inproceedings{Dellermann2019a,
   author = {Dominik Dellermann and Adrian Calma and Nikolaus Lipusch and Thorsten Weber and Sascha Weigel and Philipp Ebel},
   title = {{The Future of Human-AI Collaboration: A Taxonomy of Design Knowledge for Hybrid Intelligence Systems}},
   year = {2019},
   booktitle={Proceedings of the Fifty-Second Hawaii International Conference on System Sciences (HICSS)}, 
   publisher={University of Hawaii at Manoa, AIS, IEEE Computer Society Press},
   volume={}, 
   pages={2611--2620}
}

@article{Dellermann2019,
   author = {Dominik Dellermann and Philipp Ebel and Matthias S\"ollner and Jan Marco Leimeister},
   number = {5},
   journal = {Business and Information Systems Engineering},
   pages = {637--643},
   publisher = {Gabler Verlag},
   title = {{Hybrid Intelligence}},
   volume = {61},
   year = {2019},
}

@article{Geng2021,
   author = {Chuanxing Geng and Sheng Jun Huang and Songcan Chen},
   number = {10},
   journal = {IEEE Transactions on Pattern Analysis and Machine Intelligence},
   pages = {3614--3631},
   publisher = {IEEE Computer Society},
   title = {{Recent Advances in Open Set Recognition: A Survey}},
   volume = {43},
   year = {2021},
}

@article{Gronsund2020,
   author = {Tor Gr{\o}nsund and Margunn Aanestad},
   number = {2},
   journal = {Journal of Strategic Information Systems},
   title = {{Augmenting the algorithm: Emerging human-in-the-loop work configurations}},
   volume = {29},
   note = {Article no.: 101614},
   year = {2020},
}

@article{berente2021managing,
  title={Managing artificial intelligence},
  author={Berente, Nicholas and Gu, Bin and Recker, Jan and Santhanam, Radhika},
  journal={MIS quarterly},
  volume={45},
  number={3},
  year={2021}
}

@inproceedings{Han2021,
   author = {Lei Han and Xiao Dong and Gianluca Demartini},
   title = {{Iterative Human-in-the-Loop Discovery of Unknown Unknowns in Image Datasets}},
   year = {2021},
   booktitle={Proceedings of the Ninth AAAI Conference on Human Computation and Crowdsourcing (HCOMP)},
   pages = {72--83},
   publisher={AAAI Press},
   volume = {9},
}

@inproceedings{Hsu2020,
   author = {Yen-Chang Hsu and Yilin Shen and Hongxia Jin and Zsolt Kira},
   title = {{Generalized ODIN: Detecting Out-of-distribution Image without Learning from Out-of-distribution Data}},
   year = {2020},
   booktitle = {Proceedings of the IEEE/CVF Conference on Computer Vision and Pattern Recognition (CVPR)},
   publisher={IEEE Computer Society},
   volume={},
   number={},
   pages={10951--10960},
}

@inproceedings{Huang2016,
   author = {Gao Huang and Zhuang Liu and Laurens van der Maaten and Kilian Q. Weinberger},
   title = {{Densely Connected Convolutional Networks}},
   year = {2016},
   booktitle = {Proceedings of the IEEE Conference on Computer Vision and Pattern Recognition (CVPR)},
   publisher={IEEE Computer Society},
   volume={},
   pages={4700--4708},
}

@article{Jacobs1991,
   author = {Robert A. Jacobs and Steven J. Nowlan},
   title = {{Adaptive Mixture of Local Experts}},
   year = {1991},
   journal = {Neural Computation},
   publisher={MIT Press},
   volume={3},
   number={1},
   pages={79--87},
}

@inproceedings{jakubik2022,
   author = {Johannes Jakubik and Benedikt Blumenstiel and Michael V\"ossing and Patrick Hemmer},
   title = {{Instance Selection Mechanisms for Human-in-the-Loop Systems in Few-Shot Learning}},
   year = {2022},
   booktitle = {Proceedings of the Seventeenth International Conference on Wirtschaftsinformatik (WI)},
   publisher={AIS},
   volume={6},
   pages={},
}

@inproceedings{Kamar2016,
   author = {Ece Kamar},
   title = {{Directions in Hybrid Intelligence: Complementing AI Systems with Human Intelligence}},
   year = {2016},
   booktitle = {Proceedings of the Twenty-Fifth International Joint Conference on Artificial Intelligence (IJCAI)},
   publisher={AAAI Press},
   volume={},
   number={},
   pages={4070--4073},
}

@article{Keswani2022,
   author = {Vijay Keswani and Matthew Lease and Krishnaram Kenthapadi},
   title = {{Designing Closed Human-in-the-loop Deferral Pipelines}},
   year = {2022},
   journal={arXiv preprint arXiv:2202.04718},
   publisher={},
   volume={},
   number={},
   pages={},
}

@article{wu2022toward,
  title={Toward human-in-the-loop AI: Enhancing deep reinforcement learning via real-time human guidance for autonomous driving},
  author={Wu, Jingda and Huang, Zhiyu and Hu, Zhongxu and Lv, Chen},
  journal={Engineering},
  year={2022},
  publisher={Elsevier}
}

@techreport{CIFAR-10,
    author = {Alex Krizhevsky},
    title = {{Learning Multiple Layers of Features from Tiny Images}},
    institution={University of Toronto, Toronto},
    year={2009},
}

@article{MNIST,
   title = {{Gradient-Based Learning Applied to Document Recognition}},
   author={Lecun, Y. and Bottou, L. and Bengio, Y. and Haffner, P.},
   journal={Proceedings of the IEEE}, 
   publisher = {IEEE},
   year={1998},
   volume={86},
   number={11},
   pages={2278--2324},
}

@inproceedings{Lee2018,
   author = {Kimin Lee and Kibok Lee and Honglak Lee and Jinwoo Shin},
   title = {{A Simple Unified Framework for Detecting Out-of-Distribution Samples and Adversarial Attacks}},
   year = {2018},
   booktitle={Advances in Neural Information Processing Systems (NeurIPS)}, 
   publisher = {Curran Associates},
   volume={31},
   number={},
   pages={7167--7177},
}

@inproceedings{Liang2018,
   author = {Shiyu Liang and Yixuan Li and R. Srikant},
   title = {{Enhancing The Reliability of Out-of-distribution Image Detection in Neural Networks}},
   year = {2018},
   booktitle={Proceedings of the Sixth International Conference on Learning Representations (ICLR)}, 
   publisher = {ICLR},
   volume={},
   number={},
   pages={},
}

@inproceedings{Liu2020_OOD,
   author = {Anthony Liu and Santiago Guerra and Isaac Fung and Gabriel Matute and Ece Kamar and Walter Lasecki},
   booktitle = {Proceedings of the World Wide Web Conference (WWW)},
   pages = {2432--2442},
   volume ={},
   publisher = {ACM},
   title = {{Towards Hybrid Human-AI Workflows for Unknown Unknown Detection}},
   year = {2020},
}

@inproceedings{Liu2020_EnergyOOD,
   author = {Weitang Liu and Xiaoyun Wang and John D. Owens and Yixuan Li},
   title = {{Energy-based Out-of-distribution Detection}},
   year = {2020},
   booktitle={Advances in Neural Information Processing Systems (NeurIPS)}, 
   publisher = {Curran Associates},
   volume = {33},
   pages = {21464--21475},
}

@inproceedings{Mohseni2020,
   author = {Sina Mohseni and Mandar Pitale and Jbs Yadawa and Zhangyang Wang},
   year = {2020},
   title = {{Self-Supervised Learning for Generalizable Out-of-Distribution Detection}},
   booktitle={Proceedings of the Thirty-Fourth AAAI Conference on Artificial Intelligence (AAAI-20)}, 
   publisher={AAAI Press},
   volume = {34},
   pages = {5216--5223},
}

@inproceedings{SVHN,
   author = {Yuval Netzer and Tao Wang and Adam Coates and Alessandro Bissacco and Bo Wu and Andrew Y Ng},
   title = {{Reading Digits in Natural Images with Unsupervised Feature Learning}},
   year = {2011},
   booktitle = {Workshop on Deep Learning and Unsupervised Feature Learning of the Twenty-Fifth Conference on Neural Information Processing Systems (NeurIPS)},
   publisher={Curran Associates},
}

@article{Russakovsky2014,
   author = {Olga Russakovsky and Jia Deng and Hao Su and Jonathan Krause and Sanjeev Satheesh and Sean Ma and Zhiheng Huang and Andrej Karpathy and Aditya Khosla and Michael Bernstein and Alexander C. Berg and Li Fei-Fei},
   title = {{ImageNet Large Scale Visual Recognition Challenge}},
   year = {2014},
   journal = {International Journal of Computer Vision},
   publisher = {Axel Springer},
   volume={115},
   number={},
   pages={211--252},
}

@article{Wan2021,
   author = {Sen Wan and Yimin Hou and Feng Bao and Zhiquan Ren and Yunfeng Dong and Qionghai Dai and Yue Deng},
   number = {7},
   journal = {IEEE Transactions on Neural Networks and Learning Systems},
   pages = {3287--3292},
   publisher = {Institute of Electrical and Electronics Engineers},
   title = {{Human-in-the-Loop Low-Shot Learning}},
   volume = {32},
   year = {2021},
}

@article{Wu2021_DLA,
   author = {Xingjiao Wu and Tianlong Ma and Xin Li and Qin Chen and Liang He},
   title = {{Human-In-The-Loop Document Layout Analysis}},
   year = {2021},
   journal={arXiv preprint arXiv:2108.02095},
}

@article{Wu2021,
   author = {Xingjiao Wu and Luwei Xiao and Yixuan Sun and Junhang Zhang and Tianlong Ma and Liang He},
   title = {{A Survey of Human-in-the-loop for Machine Learning}},
   year = {2021},
   journal={arXiv preprint arXiv:2108.00941},
}

@article{Fashion-MNIST,
  author       = {Han Xiao and Kashif Rasul and Roland Vollgraf},
  title        = {{Fashion-MNIST: a Novel Image Dataset for Benchmarking Machine Learning Algorithms}},
  journal={arXiv preprint arXiv:1708.07747},
  year={2017},
}

@inproceedings{Zagoruyko2016,
   author = {Sergey Zagoruyko and Nikos Komodakis},
   title = {{Wide Residual Networks}},
   year = {2016},
   booktitle = {Proceedings of the Twenty-Seventh British Machine Vision Conference (BMVC)},
   publisher = {BMVA Press},
   note={Article no.: 87},
}

@inproceedings{sanchez2022deep,
  title={Deep Learning Uncertainty in Machine Teaching},
  author={Sanchez, T{\'e}o and Caramiaux, Baptiste and Thiel, Pierre and Mackay, Wendy E},
  booktitle={27th International Conference on Intelligent User Interfaces},
  pages={173--190},
  year={2022}
}

@inproceedings{cai2019human,
  title={Human-centered tools for coping with imperfect algorithms during medical decision-making},
  author={Cai, Carrie J and Reif, Emily and Hegde, Narayan and Hipp, Jason and Kim, Been and Smilkov, Daniel and Wattenberg, Martin and Viegas, Fernanda and Corrado, Greg S and Stumpe, Martin C and others},
  booktitle={Proceedings of the 2019 CHI Conference on Human Factors in Computing Systems},
  pages={1--14},
  year={2019}
}

@inproceedings{wang2016human,
  title={Human-in-the-loop person re-identification},
  author={Wang, Hanxiao and Gong, Shaogang and Zhu, Xiatian and Xiang, Tao},
  booktitle={European conference on computer vision},
  pages={405--422},
  year={2016},
  organization={Springer}
}

@article{wrede2019smart,
  title={Smart computational exploration of stochastic gene regulatory network models using human-in-the-loop semi-supervised learning},
  author={Wrede, Fredrik and Hellander, Andreas},
  journal={Bioinformatics},
  volume={35},
  number={24},
  pages={5199--5206},
  year={2019},
  publisher={Oxford University Press}
}

@inproceedings{weber2021better,
  title={It is better to Verify: Semi-Supervised Learning with a human in the loop for large-scale NLU models},
  author={Weber, Verena and Piovano, Enrico and Bradford, Melanie},
  booktitle={Proceedings of the Second Workshop on Data Science with Human in the Loop: Language Advances},
  pages={8--15},
  year={2021}
}

@inproceedings{liu2019deep,
  title={Deep reinforcement active learning for human-in-the-loop person re-identification},
  author={Liu, Zimo and Wang, Jingya and Gong, Shaogang and Lu, Huchuan and Tao, Dacheng},
  booktitle={Proceedings of the IEEE/CVF international conference on computer vision},
  pages={6122--6131},
  year={2019}
}

@inproceedings{liang2017human,
  title={Human-in-the-loop reinforcement learning},
  author={Liang, Huanghuang and Yang, Lu and Cheng, Hong and Tu, Wenzhe and Xu, Mengjie},
  booktitle={2017 Chinese Automation Congress (CAC)},
  pages={4511--4518},
  year={2017},
  organization={IEEE}
}

@inproceedings{elmalaki2021fair,
  title={Fair-iot: Fairness-aware human-in-the-loop reinforcement learning for harnessing human variability in personalized iot},
  author={Elmalaki, Salma},
  booktitle={Proceedings of the International Conference on Internet-of-Things Design and Implementation},
  pages={119--132},
  year={2021}
}
